\documentclass[fleqn,usenatbib]{mnras}

\usepackage[T1]{fontenc}

\DeclareRobustCommand{\VAN}[3]{#2}
\let\VANthebibliography\thebibliography
\def\thebibliography{\DeclareRobustCommand{\VAN}[3]{##3}\VANthebibliography}

\usepackage{graphicx}	
\usepackage{amsmath}	
\usepackage{amssymb}	
\usepackage{fontawesome}
\usepackage[dvipsnames]{xcolor}
\usepackage{caption}
\usepackage{subcaption}
\usepackage{xcolor} 

\usepackage{newtxtext,newtxmath}

\newcommand{\au}{\,\textsc{au}}
\title[Hierarchical quadruple-star stability]{Quadruple-star systems are not always nested triples: a machine learning approach to dynamical stability}

\author[Vynatheya et al.]{
Pavan Vynatheya$^{1}$, Rosemary A. Mardling$^{2}$ and Adrian S. Hamers$^{1}$
\\
$^{1}$Max-Planck-Institut für Astrophysik, Karl-Schwarzschild-Straße 1, 85748 Garching bei München, Germany \\
$^{2}$School of Physics and Astronomy, Monash University, Clayton Victoria 3800, Australia
}

\date{Accepted XXX. Received YYY; in original form ZZZ}

\pubyear{2022}

\begin{document}
\label{firstpage}
\pagerange{\pageref{firstpage}--\pageref{lastpage}}
\maketitle

\begin{abstract}
The dynamical stability of quadruple-star systems has traditionally been treated as a problem involving two `nested' triples which constitute a quadruple. In this novel study, we employed a machine learning algorithm, the multi-layer perceptron (MLP), to directly classify 2+2 and 3+1 quadruples based on their stability (or long-term boundedness). The training data sets for the classification, comprised of $5\times10^5$ quadruples each, were integrated using the highly accurate direct $N$-body code MSTAR. We also carried out a limited parameter space study of zero-inclination systems to directly compare quadruples to triples. We found that both our quadruple MLP models perform better than a `nested' triple MLP approach, which is especially significant for 3+1 quadruples. The classification accuracies for the 2+2 MLP and 3+1 MLP models are 94\% and 93\% respectively, while the scores for the `nested' triple approach are 88\% and 66\% respectively. This is a crucial implication for quadruple population synthesis studies. Our MLP models, which are very simple and almost instantaneous to implement, are available on GitHub, along with Python3 scripts to access them.
\end{abstract}


\section{Introduction} \label{sec:intro}

Recent decades have witnessed an increasing interest in the study of small-$N$ stellar systems, both from the theoretical and observational point of view. Advances in telescope technology have revealed the true stellar multiplicity of many `single' and `binary' star systems and hence the prevalence of multiple-star systems, vindicating predictions of theories of star formation such as \cite{2001AJ....122..432R} and \cite{2004MNRAS.351..617D}. A comprehensive list of all observed multiple-star systems detected to date can be found in the Multiple Star Catalog \citep{1997A&AS..124...75T,2018ApJS..235....6T}. In particular, \citet{2017ApJS..230...15M} found that over 50\% of high-mass O- and B-type stars dwell in triples and quadruples, compared to less than 10\% in the solar-mass range. These high-mass stars (which eventually become neutron stars and black holes) are crucial for many high-energy stellar phenomena including supernovae, X-ray binaries and gravitational wave events.  Thus, their study is incomplete without understanding triples and quadruples, and consequently their dynamics and long-term stability.

The study of quadruple stability is also crucial in the context of population synthesis studies of quadruples (e.g., \citealp{2021MNRAS.506.5345H,2022ApJ...926..195V}, in the context of merger of black holes and neutron stars). In such statistical studies, it is important that the initial sampling of quadruple systems is appropriately carried out. An inaccurate stability criterion can result in either a significant fraction of unstable systems (which are either short-lived or cannot form at all) being part of the sampled data set or many stable systems being left out. In other words, a poor classifier can systematically alter calculated rates of stellar events and other statistics.

\begin{figure}
	\includegraphics[width=\columnwidth]{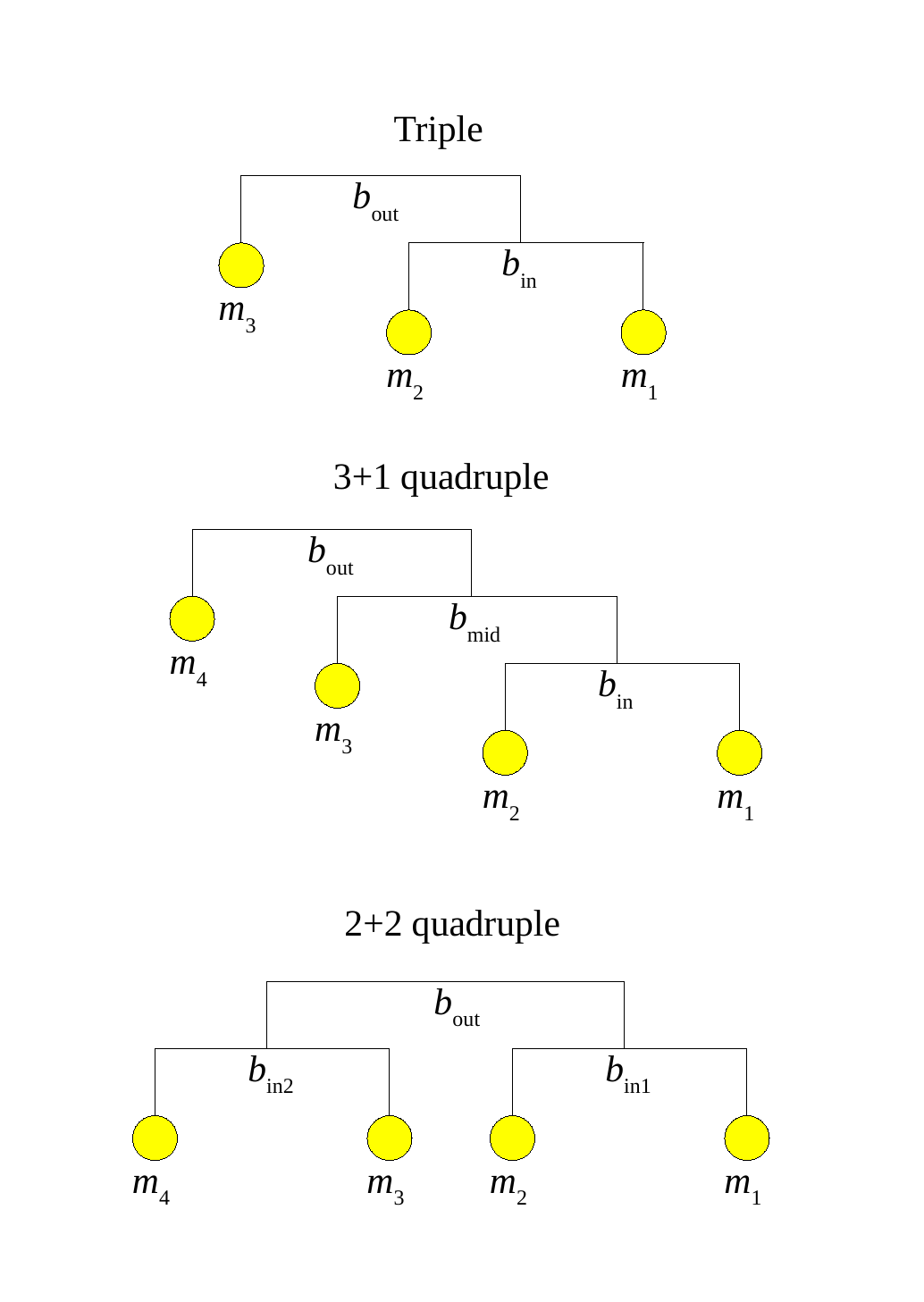}
    \caption{Mobile diagrams of a triple and the two types of quadruple-star systems (adapted from \citealp{2022ApJ...926..195V}). Here, $m_i$s refer to individual stellar masses and $b_i$s are the binaries which constitute the quadruple.}
    \label{fig:quad_type}
\end{figure}

In our previous paper (\citealp{2022MNRAS.516.4146V}; henceforth V+22), we discussed, in detail, the stability of triples and introduced two methods to classify them into `stable' and `unstable' systems. The first classifier involved an algebraic criterion, an improvement on the pre-existing stability criterion by \cite{2001MNRAS.321..398M} (henceforth MA01). The second classifier, which is more relevant to this study, was a machine learning algorithm -- a multi-layer perceptron (MLP). In this paper, we present similar MLP models for 2+2 and 3+1 quadruple-star systems (see Figure \ref{fig:quad_type} for a mobile diagram of a triple, a 2+2 and a 3+1 quadruple). Thus, this study is a sequel of V+22, with many of the details of classification being similar.

The notable triple stability criterion of MA01 (see also \citealp{Mardling1999}) is often used to empirically determine quadruple stability, by considering quadruples as `nested' triples and applying the condition twice. The `nested' triples in the two types of hierarchical quadruples are described below (see Figure \ref{fig:quad_type} for notation reference) :
\begin{itemize}
    \item 2+2 quadruples: The first triple is the system of the inner binary $b_{\mathrm{in_1}}$, with stars of masses $m_1$ and $m_2$, and a point mass approximated tertiary companion $m_3 + m_4$. Similarly, the second triple is the system of the inner binary $b_{\mathrm{in_2}}$, with stars of masses $m_3$ and $m_4$, and a point mass approximated tertiary companion $m_1 + m_2$. The outer binary in both cases is $b_{\mathrm{out}}$.
    \item 3+1 quadruples: The first triple is the system of the inner binary $b_{\mathrm{in}}$, with stars of masses $m_1$ and $m_2$, and the intermediate star $m_3$ as the tertiary companion. The outer binary in this scenario is, hence, $b_{\mathrm{mid}}$. The second triple is the system of the intermediate binary $b_{\mathrm{mid}}$, with stars of masses $m_3$ and $m_1 + m_2$ (approximated as a point mass), and the outer star $m_4$ as the tertiary companion. The outer binary is $b_{\mathrm{out}}$.
\end{itemize}

 There have also been numerous studies in the past on the stability of hierarchical (e.g., \citealp{1995ApJ...455..640E,2006tbp..book.....V}) and non-hierarchical (e.g., \citealp{2019Natur.576..406S,2021PhRvX..11c1020G}) triples. More recently, \cite{2017MNRAS.466..276G} examined the effect of orbit inclinations on stability in the context of planets. \cite{2022ApJ...938...18L} trained a convolutional neural network on a limited time series of orbital parameters to predict the long-term stability of triples. \cite{2022ApJ...939...81H} conducted a detailed study on the disruption timescales of triples, rather than just classifying them as stable or unstable. \cite{2022PASA...39...62T} came up with an updated triple-stability criterion which takes into account the varying dependence on outer mass ratio. However, studies on quadruple-star systems are few in number, and no quadruple-specific stability criteria have been investigated. 

An important aspect of triple and quadruple dynamics is long-term secular evolution, which becomes important in timescales much larger than orbital timescales. For example, in triples, the lowest-order approximation of secular evolution manifests as the (von Zeipel)-Lidov-Kozai (LK) oscillations \citep{1910AN....183..345V,1962P&SS....9..719L,1962AJ.....67..591K}, which are periodic changes in the inner orbit eccentricities and mutual inclinations. Consequently, quadruples also undergo such evolution. 

\cite{2015MNRAS.449.4221H} analysed 3+1 quadruples in a secular-averaged approach and found that initially partially coplanar systems can become inclined if the ratio of the inner to the outer periods of LK oscillations $\mathcal{R}_0 = P_{\mathrm{LK,inner}}/P_{\mathrm{LK,outer}} \sim 1$. \cite{2018MNRAS.474.3547G} found that chaotic mutual inclination changes can occur in 3+1 quadruples under similar conditions, and that short-range forces, such as tides and rotation, can aid in inner eccentricity excitation. A detailed study of secular dynamics in both 2+2 and 3+1 quadruples was carried out by \cite{2017MNRAS.470.1657H}. They showed that high mutual inclinations, and consequently high inner eccentricities, can be achieved when the precession timescale of the outer angular momentum vector due to secular evolution is comparable to the LK period of the inner `nested' triple. Due to these factors, quadruple systems can have shorter disruption timescales, and hence can become more unstable, than their equivalent `nested' triples. Our study also arrives at a similar conclusion, but through a very different approach.

This paper is arranged as follows. Section \ref{sec:nbody} introduces the $N$-body code and subsequently, discusses dynamical stability. Section \ref{sec:mlp} goes into the details of the machine learning classifiers we use to determine quadruple stability. In Section \ref{sec:trqu}, we compare the physical differences between triples and quadruples by carrying out limited parameter space studies. We present the results of the classifications in Section \ref{sec:result}. Section \ref{sec:discuss} is the discussion and Section \ref{sec:conclude} concludes.

\section{\textit{N}-body code and stability} \label{sec:nbody}

For accurate integration of few-body systems with arbitrary masses we used the accurate direct $N$-body code MSTAR \citep{2020MNRAS.492.4131R}, restricting the study to Newtonian gravity. The latter has the advantage of making the problem scale-free (in the sense that only mass and length \textit{ratios} are relevant), but the disadvantage of excluding the stabilizing effect of general relativity when orbital speeds are an appreciable fraction of the speed of light, with post-Newtonian terms restricting the growth of eccentricity.

Consequently, the scale-free parameters for 2+2 quadruples can be defined as follows (same notation as in Figure \ref{fig:quad_type}):
\begin{itemize}
    \item Three mass ratios $q_{\mathrm{in_1}} = m_2/m_1 \le 1$ ($m_2 \le m_1$), $q_{\mathrm{in_2}} = m_4/m_3 \le 1$ ($m_4 \le m_3$), such that $m_1 + m_2 \ge m_3 + m_4$, and $q_{\mathrm{out}} = (m_3+m_4)/(m_1+m_2) \le 1$.
    \item Two semi-major axis ratios $\alpha_{\mathrm{in_1}-\mathrm{out}} = a_{\mathrm{in_1}}/a_{\mathrm{out}} < 1$ and $\alpha_{\mathrm{in_2}-\mathrm{out}} = a_{\mathrm{in_2}}/a_{\mathrm{out}} < 1$, where $a_{\mathrm{in_1}}$, $a_{\mathrm{in_1}}$ and $a_{\mathrm{out}}$ are the semi-major axes of the binaries $b_{\mathrm{in_1}}$, $b_{\mathrm{in_2}}$ and $b_{\mathrm{out}}$ respectively.
    \item Three orbital eccentricities $0 \le e_{\mathrm{in_1}} < 1$, $0 \le e_{\mathrm{in_2}} < 1$ and $0 \le e_{\mathrm{out}} < 1$ of binaries $b_{\mathrm{in_1}}$, $b_{\mathrm{in_2}}$ and $b_{\mathrm{out}}$ respectively.
    \item Three mutual inclinations $i_{\mathrm{in_1}-\mathrm{in_2}}$, $i_{\mathrm{in_1}-\mathrm{out}}$ and $i_{\mathrm{in_2}-\mathrm{out}}$ between the the binary pairs $b_{\mathrm{in_1}}$--$b_{\mathrm{in_2}}$, $b_{\mathrm{in_1}}$--$b_{\mathrm{out}}$ and $b_{\mathrm{in_2}}$--$b_{\mathrm{out}}$ respectively
\end{itemize}

Similarly, the scale-free parameters for 3+1 quadruples can be defined as follows (same notation as in Figure \ref{fig:quad_type}):
\begin{itemize}
    \item Three mass ratios $q_{\mathrm{in}} = m_2/m_1 \le 1$ ($m_2 \le m_1$), $q_{\mathrm{mid}} = m_3/(m_1+m_2)$ and $q_{\mathrm{out}} = m_4/(m_1+m_2+m_3)$.
    \item Two semi-major axis ratios $\alpha_{\mathrm{in}-\mathrm{mid}} = a_{\mathrm{in}}/a_{\mathrm{mid}} < 1$ and $\alpha_{\mathrm{mid}-\mathrm{out}} = a_{\mathrm{mid}}/a_{\mathrm{out}} < 1$, where $a_{\mathrm{in}}$, $a_{\mathrm{mid}}$ and $a_{\mathrm{out}}$ are the semi-major axes of the binaries $b_{\mathrm{in}}$, $b_{\mathrm{mid}}$ and $b_{\mathrm{out}}$ respectively.
    \item Three orbital eccentricities $0 \le e_{\mathrm{in}} < 1$, $0 \le e_{\mathrm{mid}} < 1$ and $0 \le e_{\mathrm{out}} < 1$ of binaries $b_{\mathrm{in}}$, $b_{\mathrm{mid}}$ and $b_{\mathrm{out}}$ respectively.
    \item Three mutual inclinations $i_{\mathrm{in}-\mathrm{mid}}$, $i_{\mathrm{in}-\mathrm{out}}$ and $i_{\mathrm{mid}-\mathrm{out}}$ between the the binary pairs $b_{\mathrm{in}}$--$b_{\mathrm{mid}}$, $b_{\mathrm{in}}$--$b_{\mathrm{out}}$ and $b_{\mathrm{mid}}$--$b_{\mathrm{out}}$ respectively
\end{itemize}

This results in a total of 11 parameters (each for 2+2 and 3+1 quadruples) on which stability can depend on. The three mutual inclinations are functions of individual orbit inclinations and longitudes of ascending node. However, non-identical combinations of the longitudes of ascending node can result in identical values of mutual inclinations. Our study does not take these degeneracies into account. It should also be noted that another set of orbital angles, the arguments of periapsis, can be important in the study of dynamical stability (see \citealp{2008LNP...760...59M}), thereby increasing the number of potentially dependent parameters to 14. Nevertheless, as in V+22, we simplified our problem by disregarding the dependence on the arguments of periapsis.

After setting up the initial parameters and the $N$-body integrator, one requires a robust stability criterion. This is crucial since the dynamical stability of a system can depend on the time scale considered, and hence, our results can differ with varying integration times. To handle this, we used a chaos theory-inspired approach to quantify stability, similar to MA01: given an initial data set of quadruple (or triple) systems, we constructed a nearly identical set of `ghost' systems with the same initial conditions, except for a tiny change in one of the parameters. We chose to increase the inner semi-major axis by $10^{-6} \au$ for this purpose. In the case of 2+2 quadruples, where there are two inner binaries, the one with the smaller total mass was chosen for the change of semi-major axis (in hindsight, it might have been better to choose the inner binary with the smallest binding energy). Subsequently, both data sets, the original and the `ghost', were run for 100 outer orbits. The justification of the choice of 100 outer orbits is similar to V+22 (also see Figure 1 of V+22) -- most unstable systems become unbound well before this time. Any system which becomes unbound within this duration was deemed unstable. However, if a system remained bound, we checked for the divergence between the original and its corresponding `ghost' system (similar to Lyapunov stability; see \citealp{2022A&A...659A..86P,2023ApJ...943...58H}). In particular, one can define a quantity dependent on time $t$:
\begin{equation}
    \delta(t) = \frac{a_{\mathrm{in, orig}}(t) - a_{\mathrm{in, ghost}}(t)}{a_{\mathrm{in, orig}}(t)}
\end{equation}
Here, $a_{\mathrm{in}}$ refers to the inner semi-major axis (which is initially slightly offset) and the subscripts refer to the original and `ghost' systems. $\delta(t)$ quantifies the relative divergence between the orbits or the degree of chaos. A bound system was deemed unstable if, at any time step, $\delta(t)$ exceeds $10^{-2}$. We chose the threshold value $10^{-2}$ after analysing the plots of $\delta(t)$ vs. $t$ for a number of stable and unstable systems. This `ghost' system stability definition is reasonable because unstable systems are also chaotic in nature. 

It is worth mentioning that V+22 used a different defining criterion for stability, where a system (also integrated for 100 outer orbits) is considered unstable when there is an escape of a body or a significant change in semi-major axes (see Section \ref{sec:discuss}). Thus, in this study, when we compared triple stability with quadruple stability, we used the `ghost' system approach for both to be consistent.

\section{Machine learning classifier} \label{sec:mlp}
In this era of computing and big data, machine learning (ML) has become an indispensable tool for classification and regression. ML algorithms `learn' from prior data to make predictions on unseen data. The specifics of our classification algorithm (to determine dynamical stability) are laid out in the following subsections.

\subsection{Data set and initial conditions}
For machine learning (see Section \ref{sec:mlp}), it is essential to have an evenly sampled parameter space. To that end, we sampled as follows (for both 2+2 and 3+1 quadruples): 
\begin{itemize}
    \item Masses were sampled log uniformly over 1 order of magnitude, such that the most massive star can be at most 10 times as massive as the least massive. More skewed masses were not considered because the integration of such systems could take longer. Mass ratios were then calculated.
    \item Semi-major axes were sampled uniformly as ratios of the outer semi-major axis (which was kept constant since the problem is scale-free).
    \item Eccentricities were sampled uniformly from 0 to 0.95. Higher eccentricities were not considered since they could result in close approaches (depending on the semi-major axes), thereby lengthening computing time.
    \item Orbital angles -- inclinations, longitudes of ascending node, arguments of periapsis -- and mean anomalies were sampled isotropically to ensure no biases. Mutual inclinations were then calculated from the inclinations and the longitudes. It should be emphasized that the other angles were \textit{not} considered as parameters for machine learning.
\end{itemize}
Moreover, we ensured that an inner orbit's apoapsis is smaller than an outer orbit's periapsis to maintain the hierarchy. However, secular evolution can result in increased inner eccentricities, which can disrupt hierarchy later on. We also restricted sampling quadruple systems `close' to the MA01 classification boundary of the two `nested' triples that constitute them. The MA01 formula is as follows:
\begin{equation}
    \frac{R_{\mathrm{p,crit}}}{a_{\mathrm{in}}} = 2.8 \left[ (1+q_{\mathrm{out}}) \frac{1+e_{\mathrm{out}}}{(1-e_{\mathrm{out}})^{1/2}} \right]^{2/5} \left( 1 - \frac{0.3 i_{\mathrm{mut}}}{\pi} \right)
    \label{eqn:MA_form}
\end{equation}
Here, $R_{\mathrm{p}} = a_{\mathrm{out}} (1-e_{\mathrm{out}})$, `out' and `in' refer to two orbits relative to each other. If $R_{\mathrm{p}} > R_{\mathrm{p,crit}}$, a triple is called MA01 stable, else MA01 unstable. A quadruple is only considered MA01 stable if both its `nested' triples are as well. In order to ensure that there were comparable numbers of stable and unstable systems for the purpose of good machine learning classification, we excluded 70\% (80\%) of sampled MA01 unstable 2+2 (3+1) quadruples since they outnumber MA01 stable quadruples for our set of initial conditions.

After sampling the data set consisting of $5\times10^5$ 2+2 and 3+1 quadruple systems each, we created an equal number of `ghost' systems as described in Section \ref{sec:nbody}. This data is further split into training (80\%) and testing data (20\%) for the classification algorithm. The training data is used to build the classifier, while the testing data is used to evaluate its performance. We constructed a similar data set (and a `ghost' data set) for triple-star systems to compare with quadruples.

For the $N$-body simulations, we limited the run time for an individual system to five hours. Any system which took longer was halted and ignored. The fractions of such systems for triples, 2+2 quadruples and 3+1 quadruples were found to be $8.3\times10^{-4}$, $1.4\times10^{-3}$ and $2.8\times10^{-2}$ respectively. Since these fractions were very small, they did not significantly affect machine learning.

\subsection{Multi-layer perceptron (MLP) - implementation}
As in V+22, we implemented the simplest form of a feed-forward artificial neural network (ANN) (\citealp{mcculloch1943logical}; see \citealp{hastie_09_elements-of.statistical-learning} for review) -- the Multi-layer perceptron (MLP) \citep{Rosenblatt58theperceptron}. We used the \texttt{scikit-learn} \citep{scikit-learn} package of Python3 for this purpose. Many of the details of the implementation of the MLP network are very similar to Section 5 of V+22. Nevertheless, for completion, we provide a summary below.

An MLP network consists of an input layer of our 11 initial parameters, multiple hidden layers with many neurons each and an output layer with a single output from 0 (`stable') to 1 (`unstable'). Firstly, the training data is passed as input. Each subsequent layer then passes information (through weights and an activation function) forward to its immediate neighbour, resulting in a single-valued output. Finally, the algorithm updates the weights (through gradient descent) to bring the predicted outputs closer to the actual outputs. This process is iterated until an optimum classification is reached. We employed the logistic activation function $\phi(x) = 1  / (1+e^{-x})$ and the Adam solver \citep{adamsolver} for gradient descent.

It is also important to note that tuning the hyper-parameters of an MLP network is crucial for a good classification. We tuned the hyper-parameters by running a grid of MLP models (coupled with cross-validating), and choosing the best-performing ones, mentioned below:
\begin{itemize}
    \item Network architecture: 4 hidden layers of 50 neurons each.
    \item Batch size (number of samples used for gradient descent): 1000.
    \item L2 regularization term (penalty term for large weights): $10^{-4}$. 
    \item Learning rate (step-size to update weights): $0.01$.
\end{itemize}

The same hyper-parameters were used for 2+2 and 3+1 quadruples, and triples, for training. Training these networks takes about 30 minutes on 64 cores of an AMD EPYC 7742 CPU. The results of machine learning are summarised in Section \ref{sec:result}.

\section{From a triple to a quadruple} \label{sec:trqu}
In addition to machine learning, which is a brute force classifier, we also wished to look at the physical differences between triple- and quadruple-star systems in a bottom-up approach. Given the large number of parameters needed to specify quadruple and triple configurations, we restricted our study to initially periapsis-aligned co-planar orbits with zero initial orbital phases. This reduces the intricate dependencies on 3 of the 11 parameters for quadruples, thereby simplifying our analysis.

To compare quadruples with triples, we started with a co-planar stable triple-star system with the following five parameters: semi-major axis ratio $\alpha_{\mathrm{tr}}$, mass ratios $q_{\mathrm{in,tr}}$ and $q_{\mathrm{out,tr}}$, and eccentricities $e_{\mathrm{in,tr}}$ and $e_{\mathrm{out,tr}}$. We then split one of the stars (in a co-planar way) into a `new' binary to form a quadruple. Thus, we get three extra parameters from the 'new' binary: semi-major axis $\alpha_{\mathrm{new}}$, mass ratio $q_{\mathrm{in,new}}$, and eccentricities $e_{\mathrm{in,new}}$. The splitting for the two types of quadruples is done as follows:
\begin{itemize}
    \item 2+2 quadruples: The outer star is split. The original triple parameters are equivalent to: $\alpha_{\mathrm{tr}} \equiv \alpha_{\mathrm{in_1}-\mathrm{out}}$, $q_{\mathrm{in,tr}} \equiv q_{\mathrm{in_1}}$, $q_{\mathrm{out,tr}} \equiv q_{\mathrm{out}}$, $e_{\mathrm{in,tr}} \equiv e_{\mathrm{in_1}}$, $e_{\mathrm{out,tr}} \equiv q_{\mathrm{out}}$. The `new' triple parameters are equivalent to: $\alpha_{\mathrm{new}} \equiv \alpha_{\mathrm{in_2}-\mathrm{out}}$, $q_{\mathrm{in,new}} \equiv q_{\mathrm{in_2}}$, $e_{\mathrm{in,new}} \equiv e_{\mathrm{in_2}}$.
    \item 3+1 quadruples: One of the inner stars is split. The original triple parameters are equivalent to: $\alpha_{\mathrm{tr}} \equiv \alpha_{\mathrm{mid}-\mathrm{out}}$, $q_{\mathrm{in,tr}} \equiv q_{\mathrm{mid}}$, $q_{\mathrm{out,tr}} \equiv q_{\mathrm{out}}$, $e_{\mathrm{in,tr}} \equiv e_{\mathrm{mid}}$, $e_{\mathrm{out,tr}} \equiv q_{\mathrm{out}}$. The `new' triple parameters are equivalent to: $\alpha_{\mathrm{new}} \equiv \alpha_{\mathrm{in}-\mathrm{mid}}$, $q_{\mathrm{in,new}} \equiv q_{\mathrm{in}}$, $e_{\mathrm{in,new}} \equiv e_{\mathrm{in}}$.
\end{itemize}

\begin{table}
	\centering
	\begin{tabular}{lllllllll}
        \hline
        Slice & $\alpha_{\mathrm{in_1}-\mathrm{out}}$ & $\alpha_{\mathrm{in_2}-\mathrm{out}}$ & $q_{\mathrm{in_1}}$ & $q_{\mathrm{in_2}}$ & $q_{\mathrm{out}}$ & $e_{\mathrm{in_1}}$ & $e_{\mathrm{in_2}}$ & $e_{\mathrm{out}}$ \\
        \hline
        Fiducial & 0.25 & V | V & 1 & 1 | V & 1 & 0 & V | 0 & 0 \\
        Low $q_{\mathrm{in_1}}$ & 0.2 & V | V & \textbf{1/9} & 1 | V & 1 & 0 & V | 0 & 0 \\
        Low $q_{\mathrm{out}}$ & 0.25 & V | V & 1 & 1 | V & \textbf{1/9} & 0 & V | 0 & 0 \\
        High $e_{\mathrm{in_1}}$ & 0.175 & V | V & 1 & 1 | V & 1 & \textbf{0.5} & V | 0 & 0 \\
        High $e_{\mathrm{out}}$ & 0.075 & V | V & 1 & 1 | V & 1 & 0 & V | 0 & \textbf{0.5} \\
        \hline
    \end{tabular}
	\caption{Parameter space slices (zero inclination) considered for 2+2 quadruples. The `Fiducial' slice has equal masses and zero eccentricities, while other slices change either the masses or the eccentricities. The parameters with values `V' are varied to make grid-sampled plots like Figure \ref{fig:2p2_scat_grid} (the `Fiducial' slices), the left (right) value being when the mass ratios (eccentricities) are varied with the semi-major axes of the `new' binary.}
	\label{tab:2p2_slices}
\end{table}

\begin{table}
	\centering
	\begin{tabular}{lllllllll}
        \hline
        Slice & $\alpha_{\mathrm{in}-\mathrm{mid}}$ & $\alpha_{\mathrm{mid}-\mathrm{out}}$ & $q_{\mathrm{in}}$ & $q_{\mathrm{mid}}$ & $q_{\mathrm{out}}$ & $e_{\mathrm{in}}$ & $e_{\mathrm{mid}}$ & $e_{\mathrm{out}}$ \\
        \hline
        Fiducial & V | V & 0.25 & 1 | V & 1/2 & 1/3 & V | 0 & 0 & 0 \\
        High $q_{\mathrm{mid}}$ & V | V & 0.175 & 1 | V & \textbf{7/2} & 1/9 & V | 0 & 0 & 0 \\
        High $q_{\mathrm{out}}$ & V | V & 0.15 & 1 | V & 1/2 & \textbf{7/3} & V | 0 & 0 & 0 \\
        Low $q_{\mathrm{mid}}$ & V | V & 0.2 & 1 | V & \textbf{1/6} & 3/7 & V | 0 & 0 & 0 \\
        Low $q_{\mathrm{out}}$ & V | V & 0.25 & 1 | V & 1/2 & \textbf{1/9} & V | 0 & 0 & 0 \\
        High $e_{\mathrm{mid}}$ & V | V & 0.175 & 1 | V & 1/2 & 1/3 & V | 0 & \textbf{0.5} & 0 \\
        High $e_{\mathrm{out}}$ & V | V & 0.075 & 1 | V & 1/2 & 1/3 & V | 0 & 0 & \textbf{0.5} \\
        \hline
    \end{tabular}
	\caption{Parameter space slices (zero inclination) similar to Table \ref{tab:2p2_slices} for 3+1 quadruples. Figure \ref{fig:3p1_scat_grid} represents the `Fiducial' slices. (It should be noted that, in our slices, $q_\mathrm{out}$ also varies when $q_\mathrm{mid}$ is varied, but not vice-versa.)}
	\label{tab:3p1_slices}
\end{table}

Subsequently, we varied the parameters of the `new' binary (namely semi-major axis, eccentricity and mass ratio) two at a time to discern their effects on stability. Meanwhile, the other parameters were kept constant. Tables \ref{tab:2p2_slices} and \ref{tab:3p1_slices} show the different parameter space slices made for 2+2 and 3+1 quadruples respectively. For each parameter space slice, we make two grid-sampled plots of stability, one for $q_{\mathrm{in,new}}$ vs. $\alpha_{\mathrm{new}}$ and another for $e_{\mathrm{in,new}}$ vs. $\alpha_{\mathrm{new}}$. We chose grid dimensions of $25\times25$ to sufficiently populate the limited parameter space slices.

Figures \ref{fig:2p2_scat_grid} and \ref{fig:3p1_scat_grid} represent these stability plots for the `Fiducial' parameter space slices of 2+2 and 3+1 quadruples respectively (see Section \ref{sec:result} for details).

\section{Results} \label{sec:result}
We trained three MLP models in total, the `2+2 MLP', the `3+1 MLP' and the `triple MLP' for 2+2 quadruples, 3+1 quadruples and triples respectively. To be clear, the `triple MLP' model differs from a very similar model presented in V+22 owing to their different defining criteria for stability. Appendix \ref{sec:model} details how to implement the models `2+2 MLP' and `3+1 MLP' in Python3. 

\subsection{MLP model performances}
To quantify how well a classification model performs, one needs to focus not only on the overall accuracy but also on individual class (either `stable' or `unstable') accuracies. Since there are only two classes, we can define four quantities: the numbers of true stable (TS), true unstable (TU), false stable (FS) and false unstable (FU) systems. True stable (unstable) systems are correctly classified as `stable' (`unstable') by the classifier, while false stable (unstable) systems are, from $N$-body simulations, actually unstable (stable) but wrongly classified as `stable' (`unstable') by the classifier. Using these 4 quantities, one can define the following:
\begin{itemize}
    \item Overall score: Total fraction of systems predicted correctly, independent of class; $\displaystyle S = T/(T+F)$, where $T = TS+TU$ and $F = FS+FU$.
    \item Precision: Fraction of predicted `stable'/`unstable' (by the classifier) systems that are actually stable/unstable (from $N$-body simulations); $P_{\mathrm{stable}} = TS/(TS+FS)$ and $P_{\mathrm{unstable}} = TU/(TU+FU)$.
    \item Recall: Fraction of actually stable/unstable systems (from $N$-body simulations) that are predicted `stable'/`unstable' (by the classifier); $R_{\mathrm{stable}} = TS/(TS+FU)$ and $R_{\mathrm{unstable}} = TU/(TU+FS)$.
\end{itemize}
The precisions and recalls, respectively, are the quantities which detail the validity and completeness of a class-wise prediction.

Tables \ref{tab:2p2_scores} and \ref{tab:3p1_scores} represent the scores, precisions and recalls of different classifiers for 2+2 and 3+1 quadruples respectively. `MA01' refers to the triple stability criterion by MA01, with the formula being applied to the `nested' triples (Section \ref{sec:intro}) that make up a quadruple. Similarly, `triple MLP' refers to that model being applied to the `nested' triples. The other two MLP models `2+2 MLP' and `3+1 MLP' are applied directly to the respective quadruples. 



\begin{table}
	\centering
	\begin{tabular}{llllll}
        \hline
        Classifier & $S$ & $P_{\mathrm{stable}}$ & $P_{\mathrm{unstable}}$ & $R_{\mathrm{stable}}$ & $R_{\mathrm{unstable}}$ \\
        \hline
        MA01 & 0.83 & 0.77 & 0.96 & 0.95 & 0.78 \\
        triple MLP & 0.88 & 0.85 & 0.94 & 0.93 & 0.87 \\
        2+2 MLP & 0.94 & 0.94 & 0.95 & 0.94 & 0.95 \\
        \hline
    \end{tabular}
	\caption{Classification results of different classifiers for 2+2 quadruples -- overall scores, precisions and recalls for truly stable and unstable systems.}
	\label{tab:2p2_scores}
\end{table}

\begin{table}
	\centering
	\begin{tabular}{llllll}
        \hline
        Classifier & $S$ & $P_{\mathrm{stable}}$ & $P_{\mathrm{unstable}}$ & $R_{\mathrm{stable}}$ & $R_{\mathrm{unstable}}$ \\
        \hline
        MA01 & 0.56 & 0.54 & 0.95 & 0.95 & 0.55 \\
        triple MLP & 0.66 & 0.59 & 0.97 & 0.96 & 0.62 \\
        3+1 MLP & 0.93 & 0.91 & 0.95 & 0.91 & 0.95 \\
        \hline
    \end{tabular}
	\caption{Classification results similar to Table \ref{tab:2p2_scores} for 3+1 quadruples.}
	\label{tab:3p1_scores}
\end{table}

The performance indicators reveal that the quadruple MLP models are significantly better in classification than the `nested' triples approach, especially for 3+1 quadruples. This is because 3+1 quadruples show the unique behaviour of the intermediate eccentricity becoming excited by the outer orbit, subsequently triggering instability of the inner triple.

The `triple MLP' model performs better than `MA01' but still falls short of good classification. The low values of the indicators $P_{\mathrm{stable}}$ (0.85 for 2+2 and 0.59 for 3+1) and $R_{\mathrm{unstable}}$ (0.87 for 2+2 and 0.62 for 3+1) tell us that the `triple MLP' (and 'MA01') model overestimates the number of stable systems. Specifically, for 3+1 quadruples, the number of false stable systems is so large that the model performs only slightly better than a random classifier.

\subsection{Analysing the stability criterion}
We also performed a sanity check of our stability criterion by investigating the boundedness of a limited sample (to keep the computational expense in check) of systems over 1000 outer orbits\footnote{We do not employ the `ghost' orbit stability criterion in this case, and instead examine the boundedness. This is because the $\delta(t)$ threshold stated in Section \ref{sec:nbody} was tailored to integration over 100 outer orbits, and not 1000.}. A random sample of 1000 (out of an initial $5\times10^6$) each of 2+2 and 3+1 quadruples compare with their assigned stability labels as follows:

\begin{itemize}
    \item 92\% and 85\% respectively of 2+2 and 3+1 quadruples which are classified `stable' remain bound after 1000 outer orbits.
    \item 92\% and 91\% respectively of 2+2 and 3+1 quadruples which are classified `unstable' become unbound within 1000 outer orbits.
\end{itemize}

To understand these numbers, we first define the LK timescale of a triple as follows (see \citealp{2015MNRAS.452.3610A,2016ARA&A..54..441N}):

\begin{equation}
    P_{\mathrm{LK}} \approx \frac{P_{\mathrm{out}}^2}{P_{\mathrm{in}}} \left(\frac{m_{\mathrm{in,tot}} + m_{\mathrm{out}}}{m_{\mathrm{out}}}\right) (1 - e_{\mathrm{out}})^{3/2}
\end{equation}

Very stable systems tend to have $P_{\mathrm{out}} \gg P_{\mathrm{in}}$ and low $e_{\mathrm{out}}$ for both `nested' triples, and hence have long LK periods. On the other hand, very unstable systems have short LK periods for at least one of the `nested' triple, in the order of 100 outer orbits. The relatively low agreement fraction for 3+1 quadruples (85\%) is primarily due to systems whose periods of LK oscillation $P_{\mathrm{LK}}$ of the two `nested' triples differ by less than one order of magnitude. We corroborated this by running another limited sample of 1000 quadruples satisfying this condition, which corresponds to the chaotic regime detailed in \cite{2017MNRAS.470.1657H} and \cite{2018MNRAS.474.3547G}. Moreover, systems with lower inclination systems have similar agreement fractions as those with higher inclination, i.e., chaotic secular evolution can happen for low initial inclinations as well \citep{2017MNRAS.470.1657H}. This fraction may reduce further if the $N$-body integration is carried out for longer. Nevertheless, for this study, the above agreement fractions are considered adequate.

\subsection{Parameter space slices}
Here, we present the results of the parameter space slices described in Section \ref{sec:trqu}. To reiterate, we restricted our study only to co-planar orbits with all orbital angles initially set to 0. Starting from a stable triple, we split one of the stars into a `new' binary to make a quadruple. We then varied the `new' binary parameters, detailed in Tables \ref{tab:2p2_slices} and \ref{tab:3p1_slices} for 2+2 and 3+1 quadruples.

\begin{figure*}
    \begin{subfigure}{\columnwidth}
        \includegraphics[width=\columnwidth]{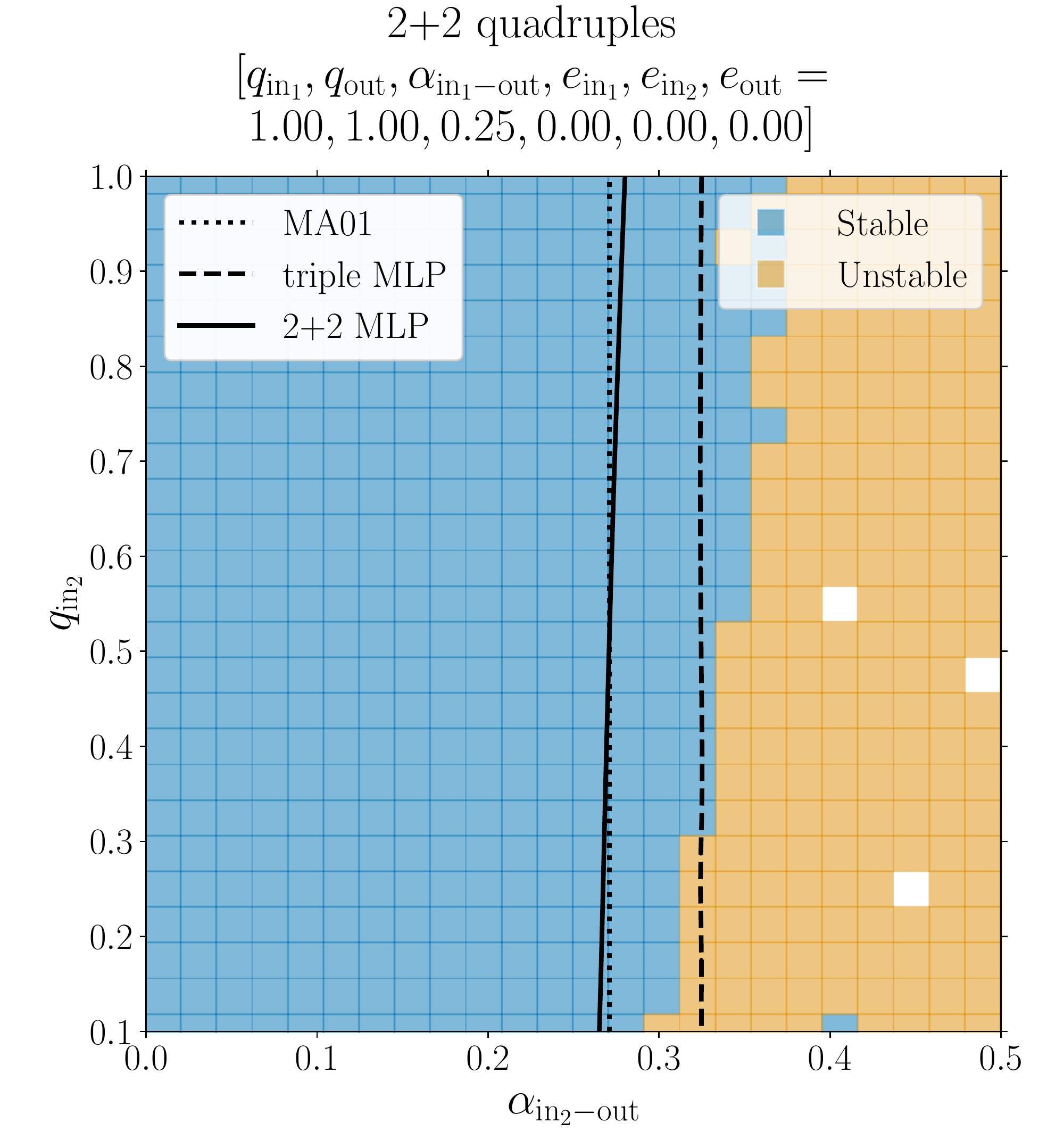}
        \caption{}
        \label{fig:2p2_scat_grid_q}
    \end{subfigure}
    \hfill
    \begin{subfigure}{\columnwidth}
        \includegraphics[width=\columnwidth]{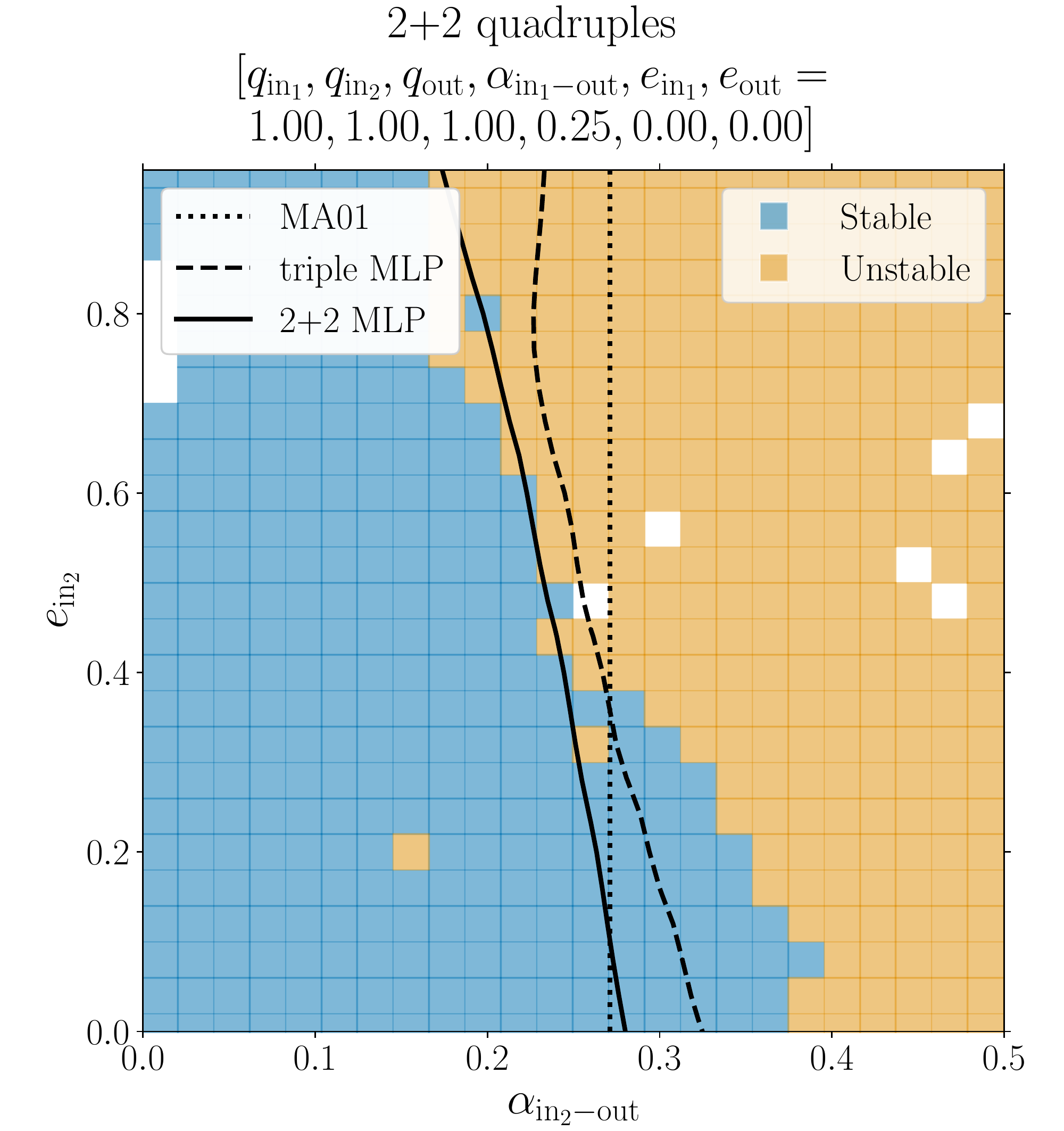}
        \caption{}
        \label{fig:2p2_scat_grid_e}
    \end{subfigure}
    \caption{Stability plots depicting the making of a 2+2 quadruple system by splitting the outer star of a stable triple system. In panel \ref{fig:2p2_scat_grid_q} (\ref{fig:2p2_scat_grid_e}), the semi-major axis ratios and the mass ratios (eccentricities) of the `new' binary are grid-sampled ($25\times25$ grid), corresponding to the `Fiducial' slices from Table \ref{tab:2p2_slices}. The constant parameters are mentioned at the top of the plots. The dotted, dashed and solid lines represent the classification boundaries as given by the three classifiers in the legend. The white spaces depict systems which took too long to run.}
    \label{fig:2p2_scat_grid}
\end{figure*}

\begin{figure*}
    \begin{subfigure}{\columnwidth}
        \includegraphics[width=\columnwidth]{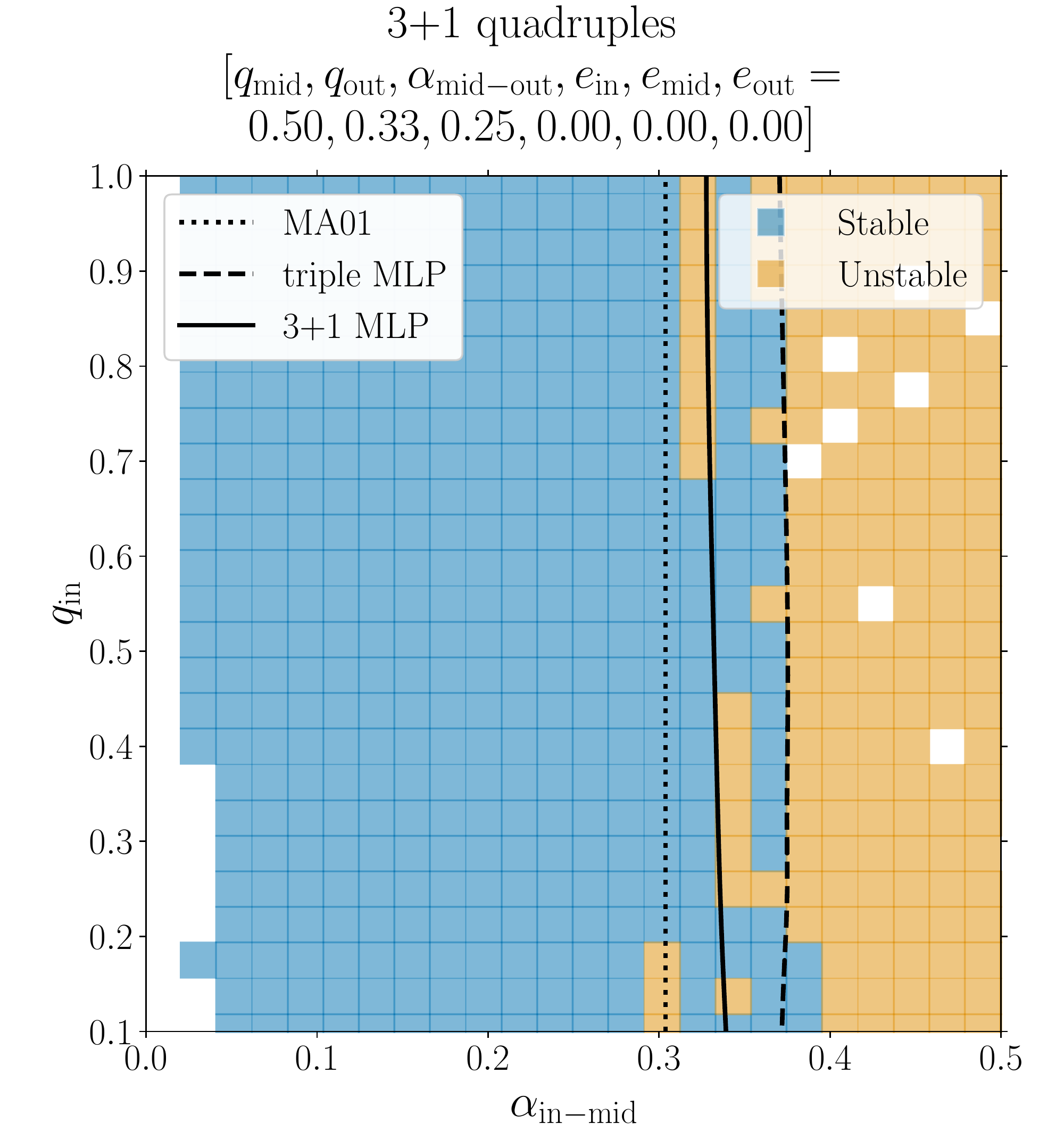}
        \caption{}
        \label{fig:3p1_scat_grid_q}
    \end{subfigure}
    \hfill
    \begin{subfigure}{\columnwidth}
        \includegraphics[width=\columnwidth]{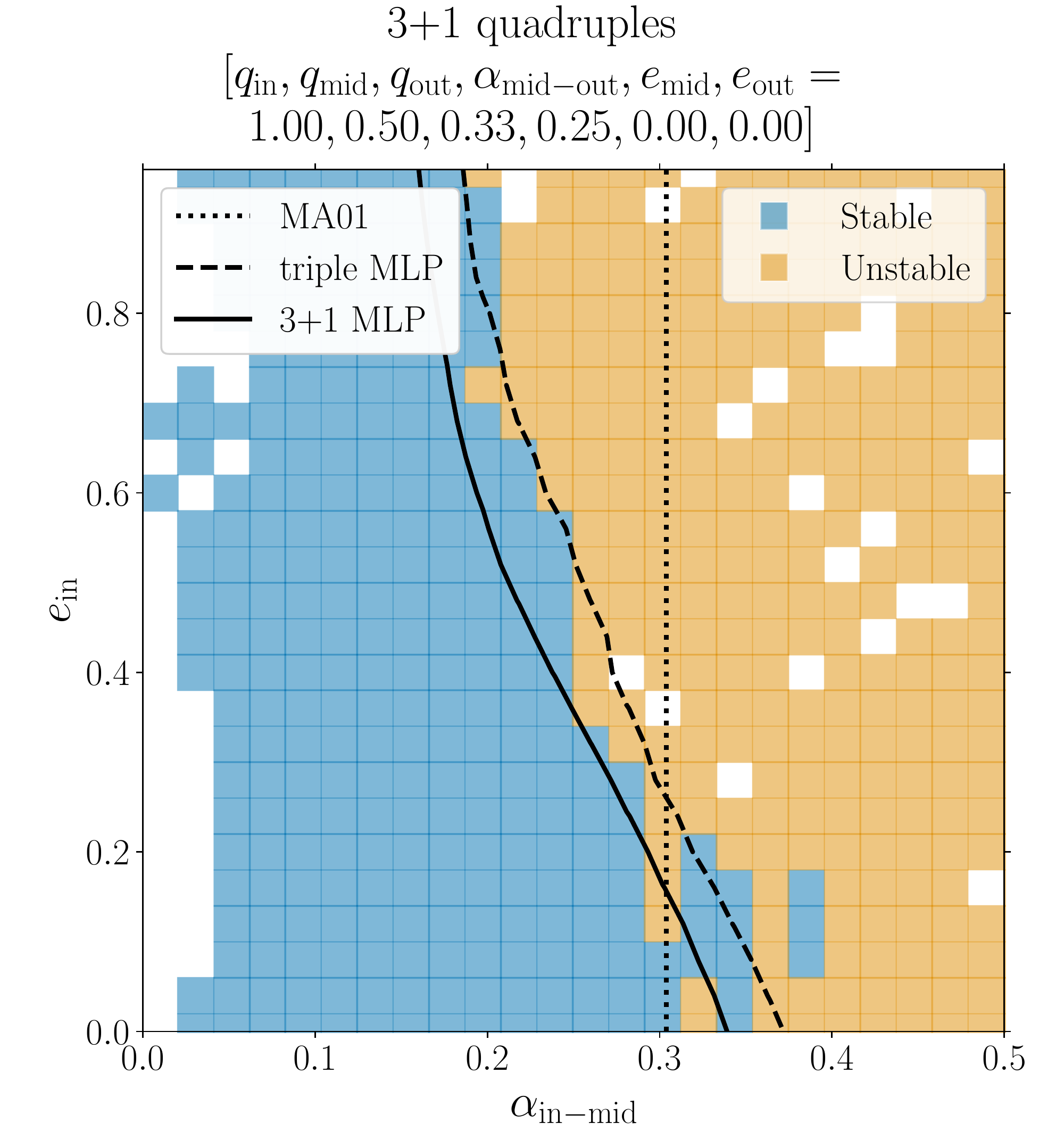}
        \caption{}
        \label{fig:3p1_scat_grid_e}
    \end{subfigure}
    \caption{Stability plots similar to \ref{fig:2p2_scat_grid} for 3+1 quadruples. These grids correspond to the `Fiducial' slices from Table \ref{tab:3p1_slices}.}
    \label{fig:3p1_scat_grid}
\end{figure*}

Figures \ref{fig:2p2_scat_grid} and \ref{fig:3p1_scat_grid} show the grid-sampled stability plots for 2+2 and 3+1 quadruples corresponding to the `Fiducial' parameter space slices in the aforementioned tables. For conciseness, the other parameter space plots are not shown, but the results are presented.

Figures \ref{fig:2p2_scat_grid_q} (\ref{fig:3p1_scat_grid_q}) and \ref{fig:2p2_scat_grid_e} (\ref{fig:3p1_scat_grid_e}) represent the plots of $q_{\mathrm{in,new}}$ vs. $\alpha_{\mathrm{new}}$ and $e_{\mathrm{in,new}}$ vs. $\alpha_{\mathrm{new}}$ respectively. The blue and orange points correspond to the systems designated stable and unstable, respectively, from the $N$-body simulations. The dotted, dashed and solid lines represent the classification boundaries of the previously-described models `MA01', `triple MLP' and `2+2 MLP'/`3+1 MLP' respectively. The white spaces correspond to the systems which took longer than five hours to run.

\begin{figure*}
    \begin{subfigure}{\columnwidth}
        \includegraphics[width=\columnwidth]{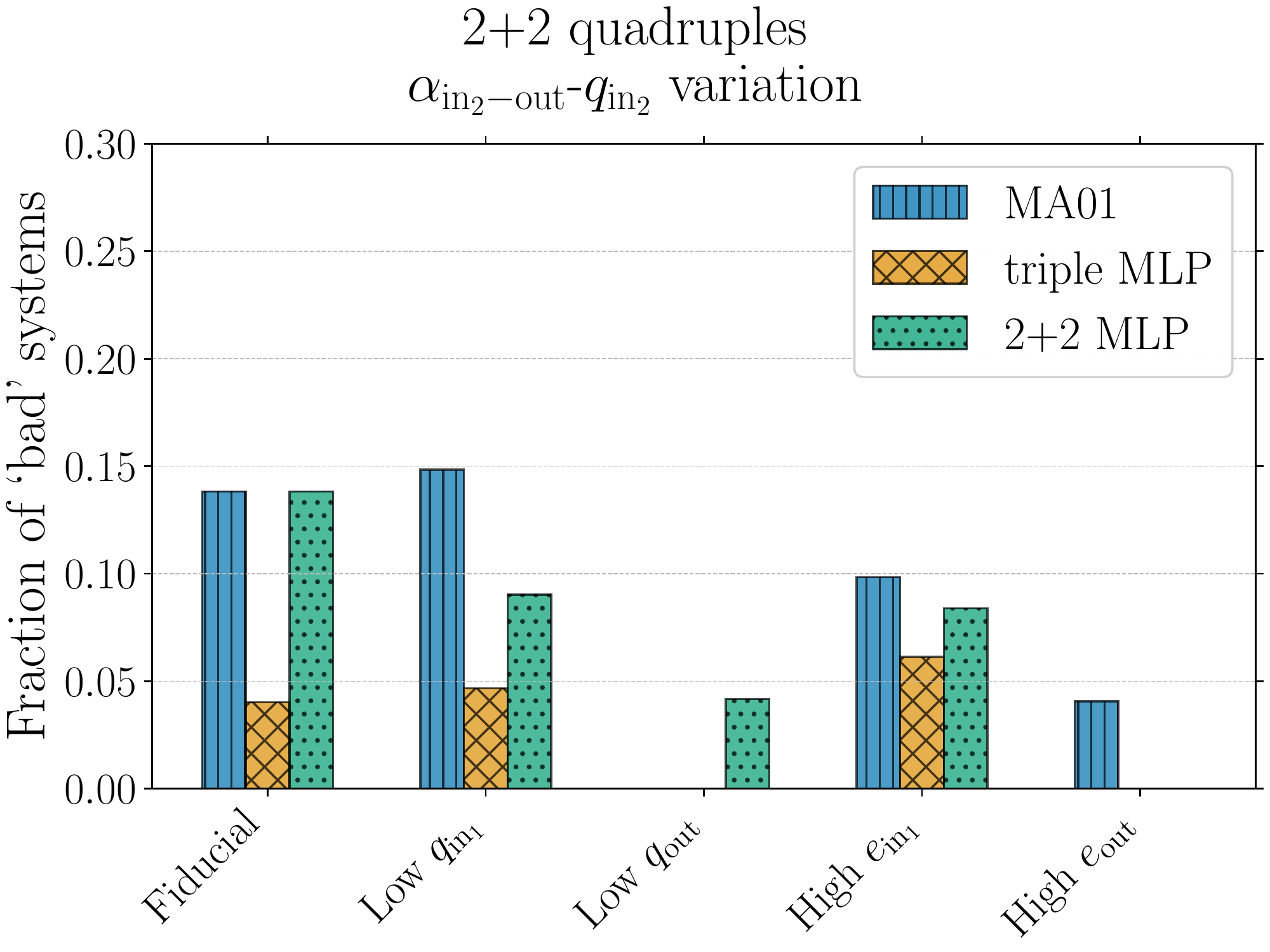}
        \caption{}
        \label{fig:2p2_bad_bar_q}
    \end{subfigure}
    \hfill
    \begin{subfigure}{\columnwidth}
        \includegraphics[width=\columnwidth]{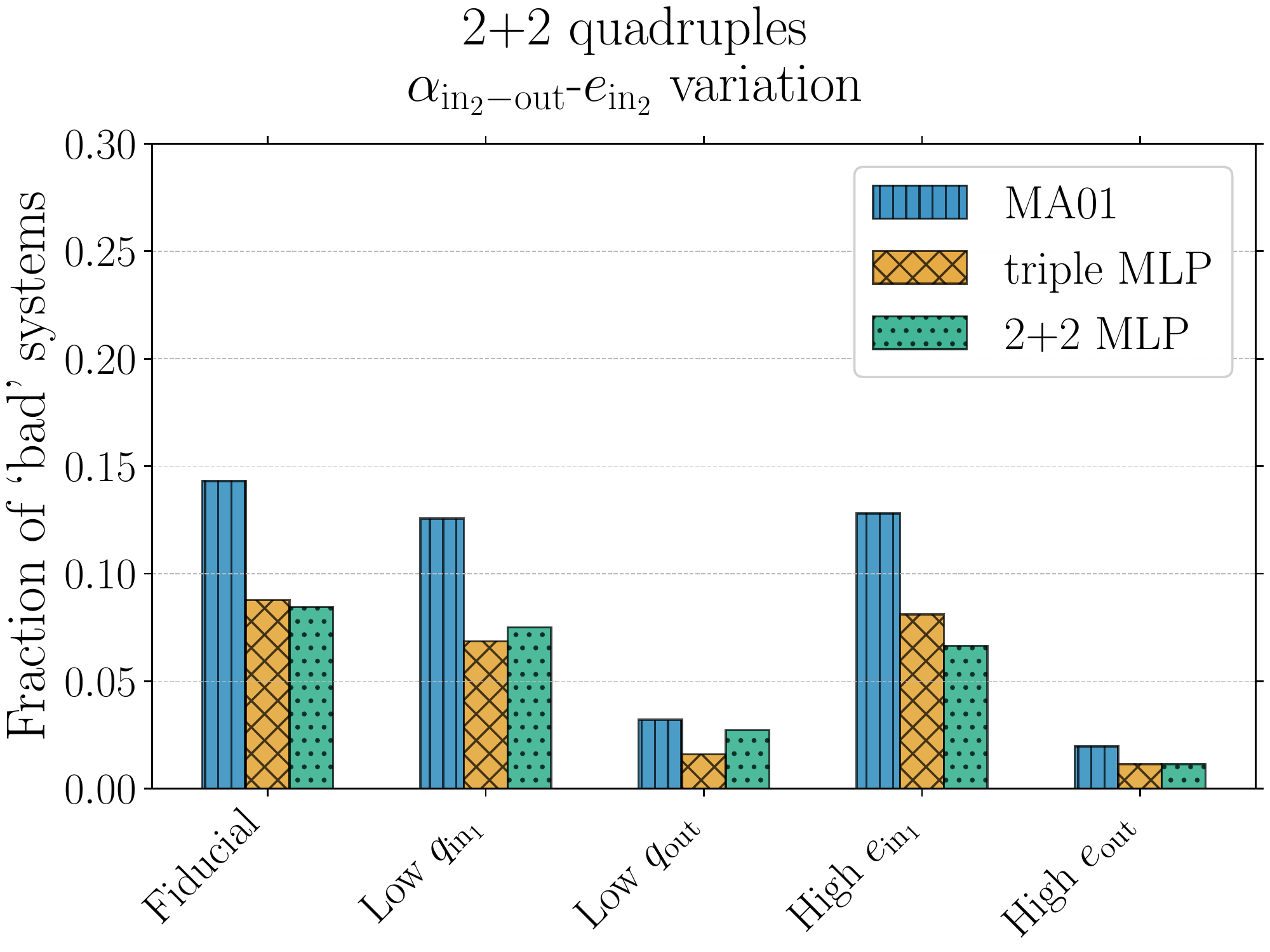}
        \caption{}
        \label{fig:2p2_bad_bar_e}
    \end{subfigure}
    \caption{Bar plots of the fraction of 2+2 quadruple systems wrongly classified (`bad') by different classifiers in different zero inclination parameter space slices, like in Figure \ref{fig:2p2_scat_grid}. In panel \ref{fig:2p2_bad_bar_q} (\ref{fig:2p2_bad_bar_e}), the semi-major axis ratios and the mass ratios (eccentricities) of the `new' binary are varied. In cases where the bars are not visible (left panel), the corresponding fractions happen to be 0. The detailed parameter values for each of the 7 parameter space slices are given in Table \ref{tab:2p2_slices}.}
    \label{fig:2p2_bad_bar}
\end{figure*}

\begin{figure*}
    \begin{subfigure}{\columnwidth}
        \includegraphics[width=\columnwidth]{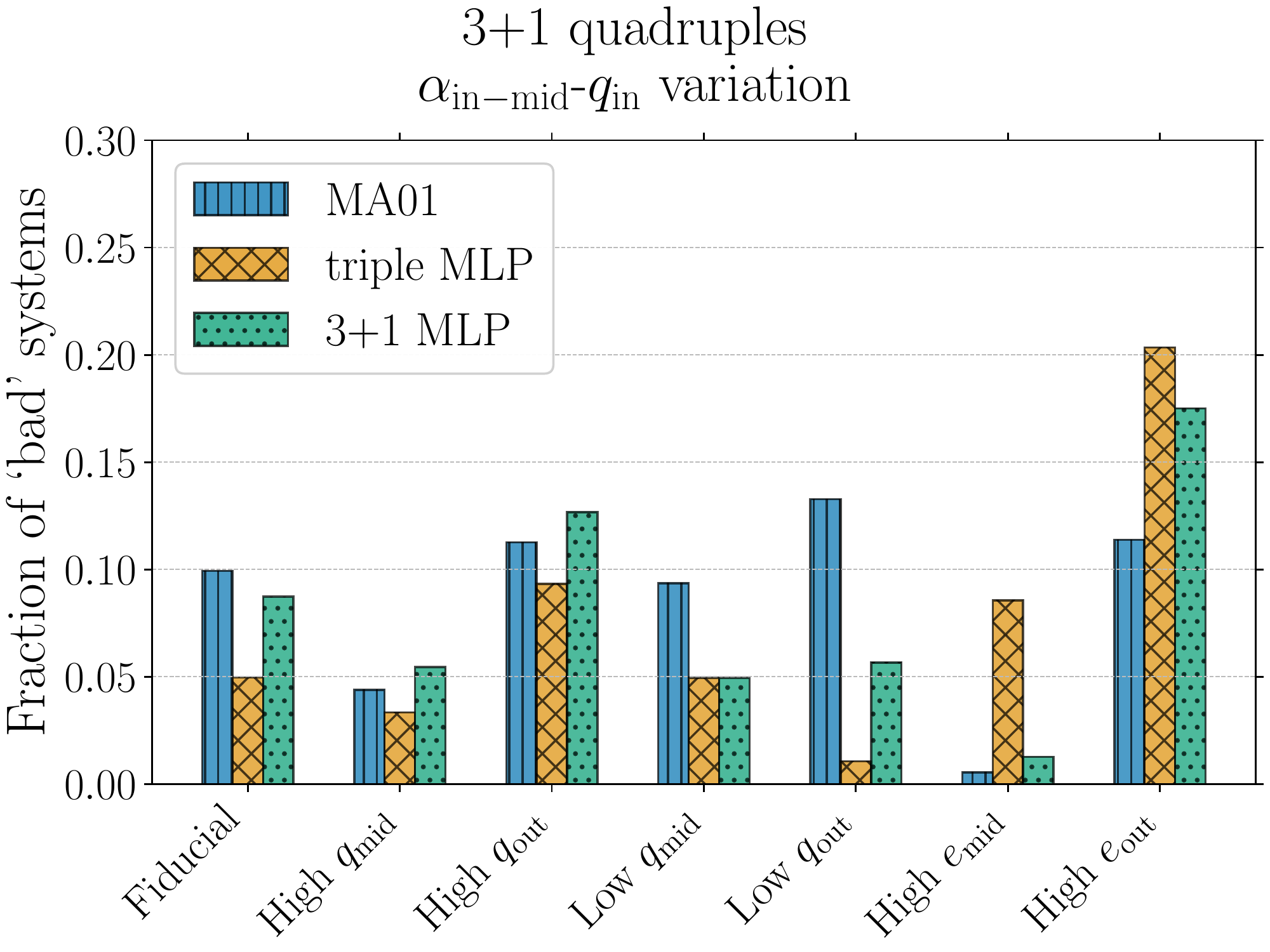}
        \caption{}
        \label{fig:3p1_bad_bar_q}
    \end{subfigure}
    \hfill
    \begin{subfigure}{\columnwidth}
        \includegraphics[width=\columnwidth]{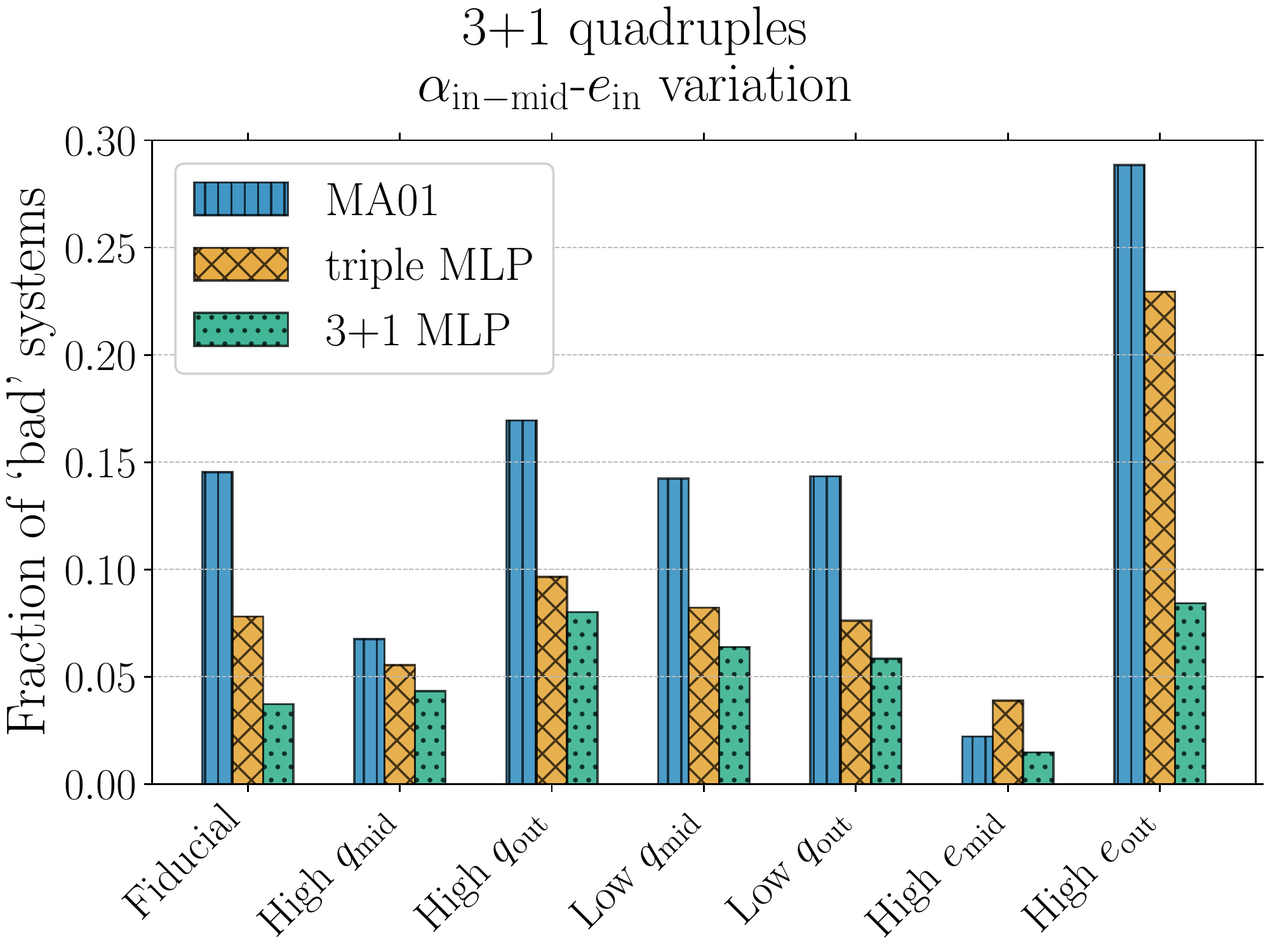}
        \caption{}
        \label{fig:3p1_bad_bar_e}
    \end{subfigure}
    \caption{Bar plots similar to \ref{fig:2p2_bad_bar} for 3+1 quadruples. The detailed parameter values for each of the 7 parameter space slices are given in Table \ref{tab:3p1_slices}.}
    \label{fig:3p1_bad_bar}
\end{figure*}

The plots indicate that the MLP classifiers perform satisfactorily, while `MA01' predicts worse when $e_{\mathrm{in,new}}$ is varied. This is evident when the fractions of wrongly classified (`bad') systems are compared. Figures \ref{fig:2p2_bad_bar} and \ref{fig:3p1_bad_bar} represent the bar charts, for 2+2 and 3+1 quadruples, of these fractions in the different parameter space slices.
\begin{itemize}
    \item For 2+2 quadruples, the fractions of `bad' systems remain lower than 15\% using any of the three classifiers. `MA01' predicts slightly worse than the others (which are comparable) when $e_{\mathrm{in_2}}$ is varied. This trend is not obvious when $q_{\mathrm{in_2}}$ is varied, although `triple' MLP performs better than the other two.
    \item For 3+1 quadruples, the fractions of `bad' systems are higher, even beyond 25\% when $e_{\mathrm{out}}$ is high, using classifier `MA01'. Again, `MA01' performs the worst when $e_{\mathrm{in}}$ is varied, while `3+1 MLP' preforms the best. All classifiers predict badly when $e_{\mathrm{out}}$ is high.
\end{itemize}
It is apparent, even from this restricted study, that the `nested' triple approximation does not work well for 3+1 quadruples, although the differences are less drastic in the co-planar case. When inclinations are included, other effects of eccentricity enhancements due to changes in mutual inclinations come into play as well, further complicating the problem.

\subsection{The complete parameter space}
\begin{figure*}
	\includegraphics[width=2.0\columnwidth]{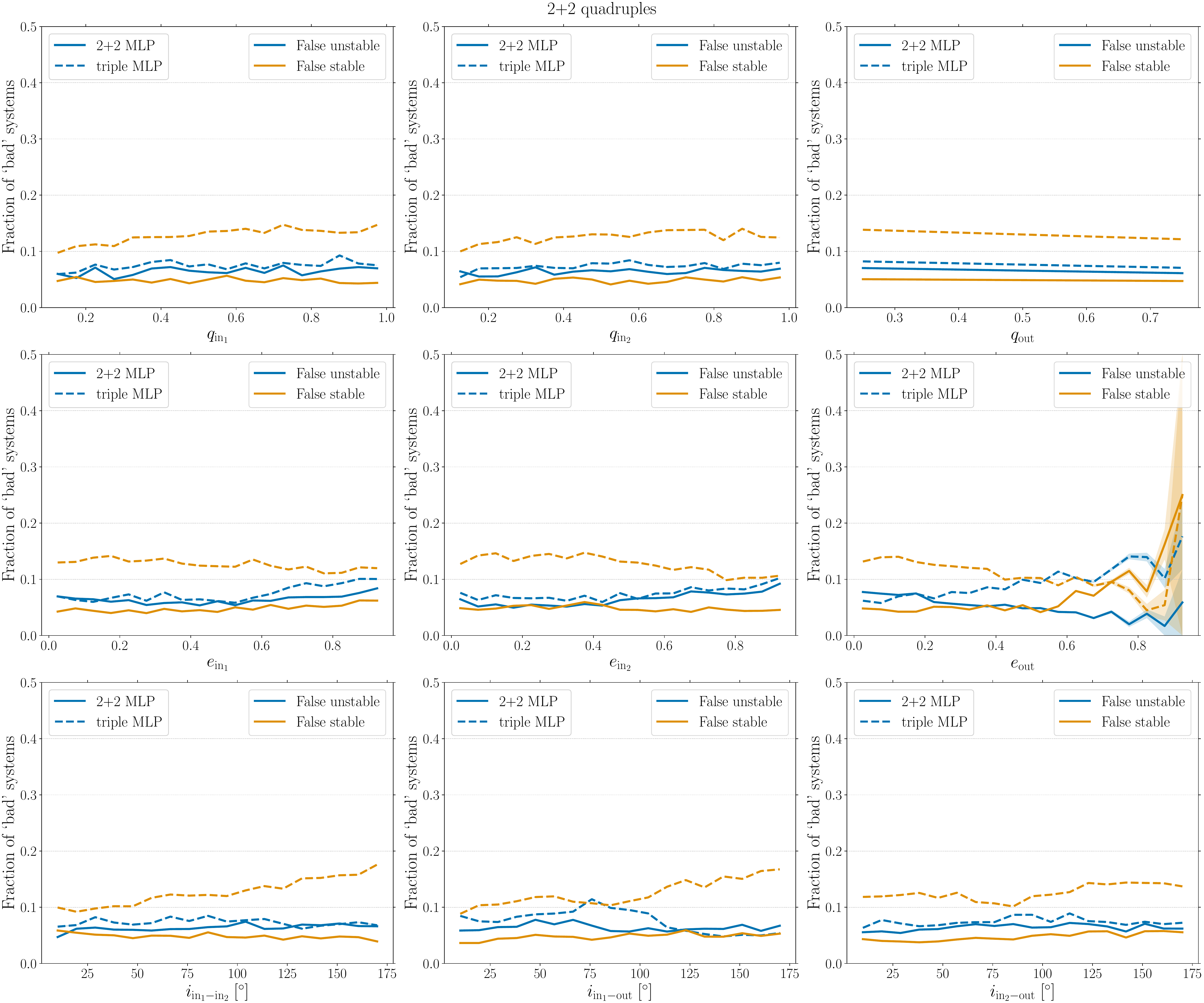}
    \caption{Frequency polygons of the fraction of 2+2 quadruple systems in the test data set wrongly classified (`bad') by the two MLP models `2+2 MLP' (solid) and 'triple MLP' (dashed). The X-axis in each panel corresponds to the considered range of each parameter. The shaded regions (most visible in the centre-right panel in the range of high $e_{\mathrm{out}}$) depict the uncertainty in the fractions.}
    \label{fig:2p2_bad_frac}
\end{figure*}

\begin{figure*}
	\includegraphics[width=2.0\columnwidth]{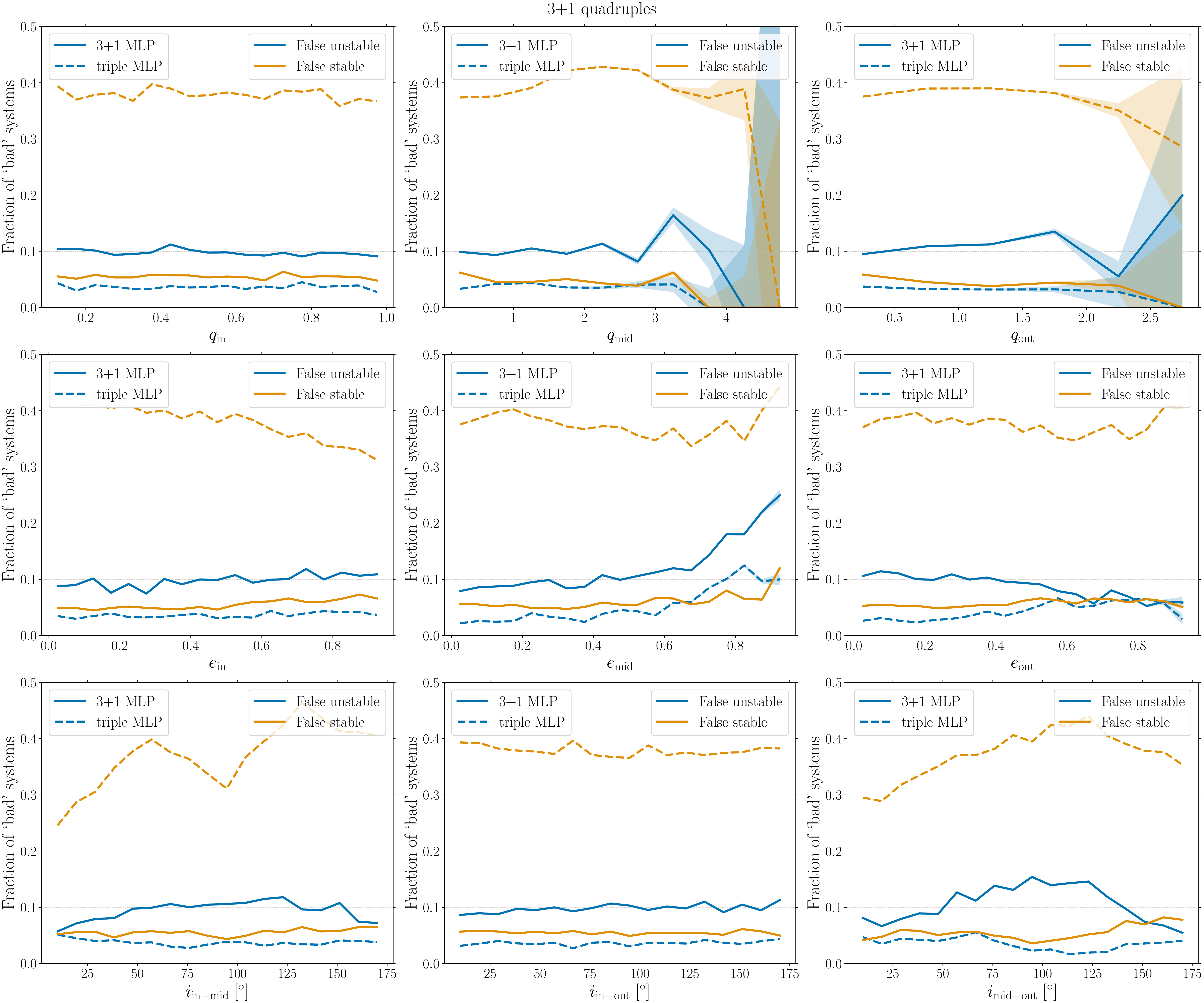}
    \caption{Frequency polygons similar to \ref{fig:2p2_bad_frac} for 3+1 quadruples. In this case, the shaded uncertainty regions are present in the ranges of high $q_{\textrm{mid}}$ (top-centre) and $q_{\textrm{out}}$ (top-right).}
    \label{fig:3p1_bad_frac}
\end{figure*}

In this section, we take a look at the complete parameter space, albeit in a simplified manner. Figure \ref{fig:2p2_bad_frac} and \ref{fig:3p1_bad_frac} display frequency polygons of the fractions of wrongly classified (`bad') systems as functions of the initial parameters of 2+2 and 3+1 quadruples respectively. In these plots, the dotted and solid lines represent the `triple MLP' and `2+2 MLP'/`3+1 MLP' models respectively. `MA01' is not depicted since it performs worse than `triple MLP'. The blue and orange lines correspond to false unstable (predicted `stable', but actually unstable) and false stable (predicted `unstable', but actually stable) systems respectively. The shaded regions represent the uncertainties in the fractions due to the relatively small number of systems being sampled in those ranges.
\begin{itemize}
    \item For 2+2 quadruples, the fractions of false unstable systems using `triple MLP' are higher than the others throughout the parameter space. In some cases, like for retrograde $i_{\mathrm{in_1}-\mathrm{in_2}}$, the false unstable fraction reaches almost 20\%. Meanwhile, the fractions for `2+2 MLP' remain lower than 10\% for most parameter ranges. In the plot with varying $e_{\mathrm{out}}$ (centre-right), there is high uncertainty in the range of large $e_{\mathrm{out}}$ due to insufficient sampling, owing to their propensity of being unstable. 
    \item For 3+1 quadruples, the fractions of false unstable systems using `triple MLP' are extremely high, around 40\%, throughout the parameter space. This result agrees with the previous sections. `3+1 MLP' has significantly lower fractions, close to 10\%, although they are higher than in the 2+2 quadruple case. The plots show that `3+1 MLP' has some trouble in classifying systems with high $e_{\mathrm{mid}}$ (again due to insufficient sampling) and near-perpendicular $i_{\mathrm{mid}-\mathrm{out}}$, where errors can reach up to 15\%. Large uncertainties, in this case, are seen in the ranges of high $q_{\mathrm{mid}}$ and $q_{\mathrm{out}}$ since high mass ratio systems are improbable when individual masses range only 1 order of magnitude.
\end{itemize}
Looking at the fraction of `bad' systems in the complete parameter space helps to identify the limitations of our models `2+2 MLP' and `3+1 MLP'. 

In summary, the `nested' triples approximation works to some extent for 2+2 quadruples but fails for 3+1 quadruples. This also makes physical sense: for 2+2 quadruples, the point-mass approximation is carried out for the two inner binaries which are relatively tight, whereas for 3+1 quadruples, one of the approximated triples ignores the influence of the outer star. This stresses the importance of secular evolution in quadruples, as detailed by \cite{2017MNRAS.470.1657H} (also see Section \ref{sec:intro}).

\section{Discussion} \label{sec:discuss}

As mentioned in Section \ref{sec:intro}, this paper should be viewed as a follow-up of V+22, which improved on the existing MA01 stability criterion and also provided a machine learning MLP classifier. However, the key difference between this study and V+22 lies in the defining criterion for stability. In V+22, a triple which becomes unbound before 100 outer orbits is deemed unstable, but bound systems face another test -- if there is a change of over 10\% in either of the two semi-major axes during any time step, the triple is called unstable, else stable (see V+22 for details).

To check how our new defining criterion, involving `ghost' systems, compared with the definition by V+22, we ran limited parameter space runs to visually see the differences. In particular, we varied two triple parameters, the semi-major axis ratio $\alpha$ and the mutual inclination $i_{\mathrm{mut}}$, keeping the others constant (similar to Figures 4 and 5 of V+22). Even with our new defining criterion, the `bump' of unstable systems for highly-inclined systems was observed, lending credence to both defining criteria. Moreover, the stability boundaries using both were nearly identical, except for highly retrograde systems ($i_{\mathrm{mut}} \gtrsim 160^{\circ}$). V+22 predicted very few stable systems when $\alpha \gtrsim 0.5$) values as compared to the new criterion, which predicts significant numbers of stable retrograde systems up to almost $\alpha \sim 0.7$. The stability of highly retrograde triples was the only ambiguity we noticed between the two defining criteria for stability. This, in turn, implies that our MLP models may not be very reliable for such systems.

In Section \ref{sec:trqu}, we highlighted that our limited parameter space study was restricted to systems with 0 initial mutual inclinations. This is because, unlike in triples where there is only one mutual inclination parameter, quadruples have three. Our preliminary study of varying mutual inclinations indicated that the stability boundary is not very well-defined in certain regions of the parameter space, possibly corresponding to resonances between different timescales. This is due to the chaotic evolution of mutual inclinations as detailed in \cite{2015MNRAS.449.4221H} and \cite{2017MNRAS.470.1657H}. The details of the resonances and the effect of mutual inclinations on stability are beyond the purview of this paper.

Finally, we note that, unlike V+22, we provide no analytical formula for the stability of quadruples. While this was initially one of the aims of this study, it has proven to be a significant challenge, especially due to the aforementioned intricate dependencies on all three mutual inclinations. Nevertheless, our `2+2 MLP' and `3+1 MLP' models can easily be implemented (see Appendix \ref{sec:model}) in population synthesis studies to weed out unstable systems more efficiently than any other method to date.

\section{Conclusion} \label{sec:conclude}
We constructed efficient machine learning models -- multi-layer perceptrons (MLPs) -- to classify quadruple-star (2+2 and 3+1) systems based on their dynamical stability. For this purpose, $5\times10^5$ 2+2 and 3+1 quadruples were generated as the training data set, and they were integrated for 100 outer orbits using the direct $N$-body code MSTAR \citep{2020MNRAS.492.4131R}. We compared the performances of `2+2 MLP' and `3+1 MLP' with a similar `triple MLP' model, which was trained on $5\times10^5$ triple systems and applied on the two `nested' triples that constitute each quadruple system. We also conducted a limited parameter space study of co-planar quadruples, to compare them directly with triples in a bottom-up approach. We started with stable triple-star systems with varying initial conditions and split one of the stars to form quadruples. The important conclusions from this paper are as follows:

\begin{itemize}
    \item The `2+2 MLP' model, a neural network of 4 hidden layers of 50 neurons each, has an overall classification score of 94\% on the testing data set. The precisions and recalls of stable (unstable) systems are 94\% (95\%) and 94\% (95\%) respectively. This is an improvement on the `triple MLP' model with a score of 88\%.
    \item The `3+1 MLP' model, also a neural network of 4 hidden layers of 50 neurons each, has an overall classification score of 93\% on the testing data set. The precisions and recalls of stable (unstable) systems are 91\% (95\%) and 91\% (95\%) respectively. This is significantly better than the `triple MLP' model with a score of just 66\%, which is only slightly better than a random classifier.
    \item For 2+2 quadruples, both `triple MLP' and 2+2 MLP' performed similarly in separating stable and unstable systems in all initially co-planar parameter space slices. The fraction of wrongly classified systems remained lower than 15\%.
    \item For 3+1 quadruples, `3+1 MLP' performed better than 'triple MLP' in all initially co-planar parameter space slices. However, both models performed badly for systems with high outer eccentricity $e_{\mathrm{out}}$.
    \item The differences in classification performance between the triple and quadruple models is less drastic for co-planar systems, which implies that mutual inclination between orbits is a significant influence on stability. 
    \item While 2+2 quadruples can still be approximated to `nested' triples up to some extent, the same approximation fails for 3+1 quadruples. In general, quadruples tend to be more unstable than their corresponding `nested' triples. This is crucial for population synthesis studies of quadruples which make use of this approximation.
    \item Our MLP models for 2+2 and 3+1 quadruples are publicly available on Github \href{https://github.com/pavanvyn/quadruple-stability}{\faGithub} in the form of a simple Python script. It is important to note that the initial parameter ranges mentioned in Section \ref{sec:mlp} need to be taken into account while using our models.  
\end{itemize}

\section*{Acknowledgements}
We thank the anonymous referee for helpful and insightful comments. A. S. H. thanks the Max Planck Society for support through a Max Planck Research Group.

\section*{Data Availability}
The data underlying this article will be shared upon reasonable request to the corresponding author.

\bibliographystyle{mnras}
\bibliography{quad_stable} 

\appendix

\section{Using our MLP models} \label{sec:model}
Our `2+2 MLP' and `3+1' MLP networks and accompanying Python3 scripts have been uploaded on GitHub \href{https://github.com/pavanvyn/quadruple-stability}{\faGithub} to ensure easy access.

The only non-basic package require to use the code is scikit-learn, which can be installed using the following terminal command:
\begin{verbatim}
pip3 install scikit-learn
\end{verbatim}

To classify 2+2 quadruples, a sample \verb |python3| terminal command is as follows:
\begin{verbatim}
python3 classify_quad_2p2.py -qi1 1.0 -qi2 1.0
        -qo 1.0 -ali1o 0.2 -ali2o 0.2 -ei1 0.0
        -ei2 0.0 -eo 0.0 -ii1i2 0.0 -ii1o 0.0
        -ii2o 0.0
\end{verbatim}
Here, the arguments \verb |qi1| , \verb |qi2| , \verb |qo| , \verb |ali1o| , \verb |ali2o| , \verb |ei1| , \verb |ei2| , \verb |eo| , \verb |ii1i2| , \verb |ii1o| and \verb |ii2o| refer to $q_{\mathrm{in_1}}$, $q_{\mathrm{in_2}}$, $q_{\mathrm{out}}$, $\alpha_{\mathrm{in_1}-\mathrm{out}}$, $\alpha_{\mathrm{in_2}-\mathrm{out}}$, $e_{\mathrm{in_1}}$, $e_{\mathrm{in_2}}$, $e_{\mathrm{out}}$, $i_{\mathrm{in_1}-\mathrm{in_2}}$, $i_{\mathrm{in_1}-\mathrm{out}}$ and $i_{\mathrm{in_2}-\mathrm{out}}$ respectively. \\

To classify 3+1 quadruples, a sample \verb |python3| terminal command is as follows:
\begin{verbatim}
python3 classify_quad_3p1.py -qi 1.0 -qm 0.5
        -qo 0.33 -alim 0.2 -almo 0.2 -ei 0.0
        -em 0.0 -eo 0.0 -iim 0.0 -iio 0.0
        -imo 0.0
\end{verbatim}
Here, the arguments \verb |qi| , \verb |qm| , \verb |qo| , \verb |alio| , \verb |almo| , \verb |ei| , \verb |em| , \verb |eo| , \verb |iim| , \verb |iio| and \verb |imo| refer to $q_{\mathrm{in}}$, $q_{\mathrm{mid}}$, $q_{\mathrm{out}}$, $\alpha_{\mathrm{in}-\mathrm{mid}}$, $\alpha_{\mathrm{mid}-\mathrm{out}}$, $e_{\mathrm{in}}$, $e_{\mathrm{mid}}$, $e_{\mathrm{out}}$, $i_{\mathrm{in}-\mathrm{mid}}$, $i_{\mathrm{in}-\mathrm{out}}$ and $i_{\mathrm{mid}-\mathrm{out}}$ respectively.

The parameter ranges should be restricted to the values given in Section \ref{sec:mlp} for optimal results.

It is also possible to import the two MLP classifiers to a custom \verb |python3| script. The input parameters can all be floating point numbers or \verb |numpy| arrays, as shown in the sample script below:
\begin{verbatim}
import numpy as np
from classify_quad_2p2 import mlp_classifier_2p2
from classify_quad_3p1 import mlp_classifier_3p1

# 2+2 quadruples: generate initial numpy arrays
# 3+1 quadruples: generate initial numpy arrays

mlp_2p2_pfile = "./mlp_model_2p2_ghost.pkl"
mlp_3p1_pfile = "./mlp_model_3p1_ghost.pkl"

mlp_2p2_stable = mlp_classifier_2p2(mlp_2p2_pfile,
        qi1, qi2, qo, ali1o, ali2o, ei1, ei2, eo,
        ii1i2, ii1o, ii2o)
mlp_3p1_stable = mlp_classifier_3p1(mlp_2p2_pfile,
        qi, qm, qo, alim, almo, ei, em, eo,
        iim, iio, imo)

# returns True if stable, False if unstable
\end{verbatim}

\bsp	
\label{lastpage}
\end{document}